\documentclass[11pt,a4paper]{article}
\usepackage[hyperref]{conll-2019}
\usepackage{times}
\usepackage{latexsym}
\usepackage{microtype}

\usepackage{amssymb}
\usepackage{tabularx}
\usepackage{booktabs}
\usepackage{graphicx}
\usepackage{multirow}
\usepackage{arydshln}
\usepackage{adjustbox}

\usepackage{url}

\aclfinalcopy 

\title{On the Importance of Subword Information for \\ Morphological Tasks in Truly Low-Resource Languages}

\author{
  Yi Zhu\textsuperscript{1}\thanks{\enspace Equal contribution, work partly done while at HITS.},
  Benjamin Heinzerling\textsuperscript{2,3$\ast$}, 
  Ivan Vuli\'c\textsuperscript{1},\\
  {\bf Michael Strube\textsuperscript{4}, Roi Reichart\textsuperscript{5}, Anna Korhonen\textsuperscript{1}}\\
  \textsuperscript{1}Language Technology Lab, University of Cambridge\\
  \textsuperscript{2}RIKEN AIP, \textsuperscript{3}Tohoku University\\
  \textsuperscript{4}Heidelberg Institute for Theoretical Studies, \textsuperscript{5}Technion, IIT\\
  {\tt \{yz568,iv250,alk23\}@cam.ac.uk}, {\tt benjamin.heinzerling@riken.jp}\\ 
  {\tt michael.strube@h-its.org}, {\tt roiri@technion.ac.il}
}
\date{}

\begin{document}
\maketitle
\begin{abstract}
Recent work has validated the importance of subword information for word representation learning. Since subwords increase parameter sharing ability in neural models, their value should be even more pronounced in low-data regimes. In this work, we therefore provide a comprehensive analysis focused on the usefulness of subwords for word representation learning in truly low-resource scenarios and for three representative morphological tasks: fine-grained entity typing, morphological tagging, and named entity recognition. We conduct a systematic study that spans several dimensions of comparison: \textbf{1)} \textit{type of data scarcity} which can stem from the lack of task-specific training data, or even from the lack of unannotated data required to train word embeddings, or both; \textbf{2)} \textit{language type} by working with a sample of 16 typologically diverse languages including some truly low-resource ones (e.g.\ Rusyn, Buryat, and Zulu); \textbf{3)} the choice of the \textit{subword-informed word representation method}. Our main results show that subword-informed models are universally useful across all language types, with large gains over subword-agnostic embeddings. They also suggest that the effective use of subwords largely depends on the language (type) and the task at hand, as well as on the amount of available data for training the embeddings and task-based models, where having sufficient in-task data is a more critical requirement. 


\end{abstract}

\section{Introduction and Motivation}
\label{s:introduction}
Recent studies have confirmed the usefulness of leveraging subword-level information in learning word representations \cite[\textit{inter alia}]{Peters:2018deep,Heinzerling2018BPEmbTP,Grave:2018lrec,Zhu:2019arxiv}, and in a range of tasks such as sequence tagging \cite{Lample:2016neural,Akbik:2018contextual,Devlin:2019bert}, fine-grained entity typing \cite{Zhu:2019arxiv}, neural machine translation \cite{Sennrich:2016neural,Luong:2016achieving,Lample:2018phrase,DBLP:conf/naacl/DurraniDSBN19}, or general and rare word similarity \cite{Pilehvar:2018card,Zhu:2019arxiv}. The subword-informed word representation architectures leverage the internal structure of words and assume that a word's meaning can be inferred from the meaning of its constituent (i.e., subword) parts. Instead of treating each word as an atomic unit, subword-informed neural architectures reduce data sparsity by relying on parameterization at the level of subwords \cite{Bojanowski:2017tacl,Pinter:2017emnlp,Chaudhary:2018emnlp,Kudo:2018acl}.

An increasing body of work focuses on various aspects of subword-informed representation learning such as segmentation of words into subwords and composing subword embeddings into word representations \cite[\textit{inter alia}]{Lazaridou:2013acl,Cotterell:2015morphological,Cotterell:2018tacl,Avraham:2017eacl,Vania:2017acl,Kim:2018coling,Zhang:2018coling,Zhao:2018emnlp}.\footnote{An overview of a variety of subword-informed word representation architectures and different segmentation and composition strategies is provided by \newcite{Zhu:2019arxiv}.} The increased parameter sharing ability of such models is especially relevant for learning embeddings of rare and unseen words. Therefore, the importance of subword-level knowledge should be even more pronounced in \textit{low-data regimes} for \textit{truly low-resource languages}. Yet, a systematic study focusing exactly on the usefulness of subword information in such settings is currently missing in the literature. In this work, we fill this gap by providing a comprehensive analysis of subword-informed representation learning focused on low-resource setups.

Our study centers on the following axes of comparison, focusing on three representative tasks where subword-level information can guide learning, namely fine-grained entity typing (\textsc{fget}), morphological tagging (\textsc{mtag}), and named entity recognition (\textsc{ner}): \textbf{1)} Since data scarcity can stem from unavailability of (i) task-specific training data or (ii) unannotated corpora to train the embeddings in the first place, or (iii) both, we analyse how different data regimes affect the final task performance. \textbf{2)} We experiment with 16 languages representing 4 diverse morphological types, with a focus on truly low-resource languages such as Zulu, Rusyn, Buryat, or Bambara. \textbf{3)} We experiment with a variety of subword-informed representation architectures, where the focus is on unsupervised, widely portable language-agnostic methods such as the ones based on character n-grams \cite{Luong:2016achieving,Bojanowski:2017tacl}, Byte Pair Encodings (BPE) \cite{Sennrich:2016neural,Heinzerling2018BPEmbTP}, Morfessor \cite{Aaltodoc:http://urn.fi/URN:NBN:fi:aalto-201409292677}, or BERT-style pretraining and fine-tuning \cite{Devlin:2019bert} which relies on WordPieces \cite{Wu:2016arxiv}. We demonstrate that by tuning subword-informed models in low-resource settings we can obtain substantial gains over subword-agnostic models such as skip-gram with negative sampling \cite{DBLP:conf/nips/MikolovSCCD13} across the board.

The main goal of this study is to identify viable and effective subword-informed approaches for truly low-resource languages and offer modeling guidance in relation to the target task, the language at hand, and the (un)availability of general and/or task-specific training data. As expected, our key results indicate that there is no straightforward \textit{``one-size-fits-all''} solution, although certain approaches (e.g., BPE-based or character n-grams) emerge as more robust in general. The optimal subword-informed configurations are largely task-, language-, and resource-dependent: their performance hinges on a complex interplay of the multiple factors mentioned above. For instance, we show that fine-tuning pretrained multilingual BERT \cite{Devlin:2019bert,Wu:2019beto} is a viable strategy for ``double'' low-resource settings in the \textsc{ner} and \textsc{mtag} tasks, but it fails for the \textsc{fget} task in the same setting; furthermore, its performance can be matched or surpassed by other subword-informed methods in \textsc{ner} and \textsc{mtag} as soon as they obtain sufficient embedding training data. 


\section{Methodology}
\label{s:methodology}
In what follows, we further motivate our work by analyzing two different sources of data scarcity: embedding training data (termed \textit{WE data}) and task-specific training data (termed \textit{task data}). Following that, we motivate our selection of test languages and outline the subword-informed representation methods compared in our evaluation.

\vspace{1.4mm}
\noindent \textbf{Types of Data Scarcity.} The majority of languages in the world still lack basic language technology, and progress in natural language processing is largely hindered by the lack of annotated \textit{task data} that can guide machine learning models \cite{Agic:2016tacl,ponti2018modeling}. However, many languages face another challenge: the lack of \textit{large unannotated text corpora} that can be used to induce useful general features such as word embeddings \cite{adams2017cross,Fang:2017acl,ponti2018modeling}:\footnote{For instance, as of April 2019, Wikipedia is available only in 304 out of the estimated 7,000 existing languages.} i.e.\ WE data.

The absence of data has over the recent years materialized the {\em proxy fallacy}. That is, methods tailored for low-resource languages are typically tested only by proxy, simulating low-data regimes exclusively on resource-rich languages \cite{Agic:2017eacl}. While this type of evaluation is useful for analyzing the main properties of the intended low-resource methods in controlled \textit{in vitro} conditions, a complete evaluation should also provide results on true low-resource languages \textit{in vivo}. In this paper we therefore conduct both types of evaluation. Note that in this work we still focus on low-resource languages that have at least some digital footprint (see the statistics later in Table~\ref{tb:st}), while handling zero-resource languages without any available data \cite{kornai2013digital,ponti2018modeling} is a challenge left for future work.
\begin{table*}[t]
	\centering
    \def\arraystretch{0.91}
    {\footnotesize
	\begin{tabularx}{\textwidth}{l XXXXXX XXXXXX XX XX}
		\multicolumn{1}{r}{}  & \multicolumn{6}{c}{Agglutinative} & \multicolumn{6}{c}{Fusional} & \multicolumn{2}{c}{Introflexive} & \multicolumn{2}{c}{Isolating} \\ \cmidrule(lr){2-7} \cmidrule(lr){8-13} \cmidrule(lr){14-15} \cmidrule(lr){16-17}
		\multicolumn{1}{r}{} & 
		\textsc{bm} & \textsc{bxr} & \textsc{myv} & \textsc{te} & \textsc{tr} & \textsc{zu} &
		\textsc{en} & \textsc{fo} & \textsc{ga} & \textsc{got} & \textsc{mt} & \textsc{rue} &
        \textsc{am} & \textsc{he} &
		\textsc{yo} & \textsc{zh} \\ \cmidrule(lr){2-17}
		
	    \textsc{emb} & 
		40K & 372K & 207K & 5M & 5M & 69K &
		5M & 1.6M & 4.4M & 18K & 1.5M & 282K &
        659K & 5M &
		542K & 5M\\
	    \hdashline 
		 
	    \textsc{fget} & 
		29K & 760 & 740 & 13K & 60K & 36K &
		60K & 30K & 56K & 289 & 2.7K & 1.5K &
        2.2K & 60K &
		15K & 60K \\ 
		
	    \textsc{ner} & 
	    345 & 2.4K & 2.1K & 9.9K & 167K & 425 & 
	    8.9M & 4.0K & 7.6K & 475 & 1.9K & 1.6K & 
	    1.0K & 107K & 
	    3.4K & --\\ 
		
	    \textsc{mtag} & 
	    -- & -- & -- & 1.1K & 3.7K & -- & 
	    24K & -- & -- & 3.4K & 1.1K & -- & 
	    -- & 5.2K & 
	    -- & 4.0K\\ 
		\cmidrule(lr){2-17}
		\textsc{bert} &
		& & & \checkmark & \checkmark & &
		\checkmark & & \checkmark & & & &
		& \checkmark &
		\checkmark & \checkmark\\
		\bottomrule       
	\end{tabularx}}%
    \vspace{-1.5mm}
     \caption{Overview of test languages and data availability. \textsc{emb} denotes the maximum number of tokens in corresponding Wikipedias used for training embeddings. Actual Wikipedia sizes are larger than 5M for (\textsc{te}, \textsc{tr}, \textsc{en}, \textsc{he}, \textsc{zh}), but were limited to 5M tokens in order to ensure comparable evaluation settings for data scarcity simulation experiments across different languages. \textsc{fget}, \textsc{ner}, and \textsc{mtag} rows show the number of instances for the three evaluation tasks (see \S\ref{s:tasks}): number of entity mentions for \textsc{fget}, number of sentences for \textsc{ner} and \textsc{mtag}. In \textsc{mtag}, we omit languages for which UDv2.3 provides only a test set, but no training set. The \textsc{bert} row shows the languages supported by multilingual \textsc{bert}. Languages are identified by their ISO 639-1 code.} \label{tb:st}
     \vspace{-2mm}
\end{table*}

\vspace{1.4mm}
\noindent \textbf{(Selection of) Languages.}
Both \textit{sources of data scarcity} potentially manifest in degraded task performance for low-resource languages: our goal is to analyze the extent to which these factors affect downstream tasks across morphologically diverse language types that naturally come with varying data sizes to train their respective embeddings and task-based models. Our selection of test languages is therefore guided by the following goals: \textbf{a)} following recent initiatives (e.g. in language modeling) \cite{Cotterell:2018naacl,Gerz2018on}, we aim to ensure coverage of different genealogical and typological properties; \textbf{b)} we aim to cover low-resource languages with varying amounts of available WE data and task-specific data. 

We select 16 languages in total spanning 4 broad morphological types, listed in Table~\ref{tb:st}. Among these, we chose one (relatively) high-resource language for each type: Turkish (agglutinative), English (fusional), Hebrew (introflexive), and Chinese (isolating). We use these four languages to \textit{simulate} data scarcity scenarios and run experiments where we \textit{control} the degree of data scarcity related to both embedding training data and task-related data. The remaining 12 languages are treated as test languages with varying amounts of available data (see Table~\ref{tb:st}. For instance, relying on the Wikipedia data for embedding training, Gothic (\textsc{got)} is the language from our set that contains the fewest number of word tokens in its respective Wikipedia (18K, in terms of Wikipedia size this ranks it as $273$th out of $304$ Wikipedia languages); Irish Gaelic (\textsc{ga}) with 4.4M tokens is ranked $87/304$. 

\vspace{1.4mm}
\noindent \textbf{Subword-Informed Word Representations.}
We mainly follow the framework of \citet{Zhu:2019arxiv} for the construction of subword-informed word representations; the reader is encouraged to refer to the original paper for more details.
In short, to compute the representation for a given word $w\in V$, where $V$ is the word vocabulary, the framework is based on three main components: 1) \textit{segmentation} of words into subwords, 2) \textit{interaction} between subword and position embeddings, and 3) a \textit{composition function} that yields the final word embedding from the constituent subwords. \newcite{Zhu:2019arxiv} explored a large space of possible subword-informed configurations. Based on their findings, we select a representative subset of model configurations. They can be obtained by varying the components listed in Table~\ref{tb:sub_model}.

Concretely, $w$ is first segmented into an ordered subword sequence from the subword vocabulary $S$ by a deterministic subword segmentation method.
To enable automatic language-agnostic segmentation across multiple languages, we focus on unsupervised segmentation methods: we work with Morfessor \cite{Aaltodoc:http://urn.fi/URN:NBN:fi:aalto-201409292677}, character n-grams \cite{Bojanowski:2017tacl} and BPE \cite{Gage:1994}. We use the default parameters for Morfessor, and the same $3$ to $6$ character n-gram range as \citet{Bojanowski:2017tacl}.
For BPE, the number of merge operations is a tunable hyper-parameter. It controls the segmentation ``aggressiveness'': the larger the number the more conservative the BPE segmentation is.
Following \citet{Heinzerling2018BPEmbTP}, we investigate the values $\{{1e3, 1e4, 1e5}\}$: this allows us to test varying segmentation granularity in relation to different language types.

After segmentation into subwords, each subword is represented by a vector $\mathbf{s}$ from the subword embedding matrix $\mathbf{S} \in \mathbb{R}^{|S| \times d}$, where $d$ is the dimensionality of subword embeddings.
Optionally, the word itself can be appended to the subword sequence and embedded into the subword space in order to incorporate word-level information \cite{Bojanowski:2017tacl}.
To encode subword order, $\mathbf{s}$ can be further enriched by a trainable position embedding $\mathbf{p}$. We use addition to combine subword and position embeddings, namely $\mathbf{s} := \mathbf{s} + \mathbf{p}$, which has become the de-facto standard method to encode positional information \cite{Gehring2017ConvolutionalST,NIPS2017_7181,Devlin:2019bert}.

Finally, the subword embedding sequence is passed to a composition function, which computes the final word representation. \citet{Li:2018ws} and \citet{Zhu:2019arxiv} have empirically verified that composition by simple addition, among other more complex composition functions, is a robust choice. Therefore, we use addition in all our experiments.

Similar to \citet{Bojanowski:2017tacl,Zhu:2019arxiv}, we adopt skip-gram with negative sampling \cite{DBLP:conf/nips/MikolovSCCD13} as our word-level distributional model: the target word embedding is computed by our subword-informed model, and the context word is parameterized by the (word-level) context embedding matrix $\mathbf{W}_c \in \mathbb{R}^{|V| \times d_c}$.

\begin{table}[!t]
	\centering
    \def\arraystretch{0.95}
    {\footnotesize
	\begin{tabularx}{\columnwidth}{r X r}
		Component & Option & Label\\
		\toprule
		Segmentation & Morfessor & \texttt{ morf}\\
		& \textsc{BPE} & \texttt{bpeX}\\
		& char n-gram & \texttt{charn}\\
        \cmidrule(lr){2-3}
		
		Word token & exclusion & {\it w-}\\
		& inclusion & {\it w+}\\
        \cmidrule(lr){2-3}
		
		Position embedding & exclusion & {\it p-}\\
		& additive & {\it p+}\\
        \cmidrule(lr){2-3}
		
		Composition function & addition & {\it add}\\
        \bottomrule
	\end{tabularx}}%
	\vspace{-1mm}
	\caption{Components for constructing subword-informed word representations. In the \texttt{bpeX} label $X\in\{{1e3, 1e4, 1e5}\}$ denotes the BPE vocabulary size.}
	\label{tb:sub_model}
    \vspace{-1.5mm}
\end{table}

We compare subword-informed architectures to two well-known word representation models, also captured by the general framework of \newcite{Zhu:2019arxiv}: 1) the subword-agnostic skip-gram model from the \texttt{word2vec} package \citep{DBLP:conf/nips/MikolovSCCD13} (\textsc{w2v}), and 2) \texttt{fastText} (\textsc{ft}) \cite{Bojanowski:2017tacl}. The comparison to \textsc{w2v} aims to validate the potential benefit of subword-informed word representations for truly low-resource languages, while the comparison to \textsc{ft} measures the gains that can be achieved by more sophisticated and fine-tuned subword-informed architectures.
We also compare with pretrained multilingual $BERT_{base}$ \cite{Devlin:2019bert} on the languages supported by this model.



\section{Evaluation Tasks}
\label{s:tasks}
\noindent \textbf{Fine-Grained Entity Typing.}
\textsc{fget} is cast as a sequence classification problem, where an entity mention consisting of one or more tokens (e.g.\ {\it Lincolnshire}, {\it Bill Clinton}), is mapped to one of the $112$ fine-grained entity types from the FIGER inventory \cite{Ling:2012:FER:2900728.2900742,yaghoobzadeh-schutze:2015:EMNLP,Heinzerling2018BPEmbTP}.
Since entity mentions are short token sequences and not full sentences, this semi-morphological/semantic task requires a model to rely on the subword information of individual tokens in the absence of sentence context. That is, subwords can provide evidence useful for entity type classification in the absence of context. For instance, {\it Lincolnshire} is assigned the type \texttt{/location/county} as {\it -shire} is a suffix that strongly indicates a location.
Hence, \textsc{fget} is well-suited for evaluating subword-informed representations, and can benefit from the information.

\vspace{1.4mm}
\noindent \textbf{Morphological Tagging.}
\textsc{mtag} is the task of annotating each word in a sentence with features such as part-of-speech, gender, number, tense, and case.
These features are represented as a set of key-value pairs.
For example, \emph{classified} is a finite (Fin) verb (V) in indicative (Ind) mood, third person, past tense, which is annotated with the morphological tag \{{\small\textsf{POS=V, Mood=Ind, Person=3, Tense=Past, VerbForm=Fin}}\}, and the female singular third-person possessive personal pronoun \emph{her} with the morphological tag \{{\small\textsf{Gender=Fem, Number=Sing, Person=3, Poss=Yes, PronType=Prs}}\}.


\vspace{1.4mm}
\noindent \textbf{Named Entity Recognition.}
\textsc{ner} is the task of annotating textual mentions of real-world entities with their semantic type, such as \textsf{person}, \textsf{location}, and \textsf{organization}: e.g., \emph{Barack Obama (\textsf{person}) was born in Hawaii (\textsf{location}).}

\section{Experimental Setup}
\label{s:experimental}
 \noindent \textbf{Embedding Training: WE Data.} 
For training word and subword embeddings, we rely on Wikipedia text for all 16 languages, with corresponding Wikipedia sizes listed in Table~\ref{tb:st}. For training embeddings in controlled low-resource settings with our 4 ``simulation'' languages, we sample nine data points to simulate low-resource scenarios with WE data. Specifically, we sample 10K, 20K, 50K, 100K, 200K, 500K, 1M, 2M, and 5M tokens of article text for each of the 4 languages. For the other 12 languages we report results obtained by training embedding on the full Wikipedia edition.


\vspace{1.4mm}
\noindent \textbf{Task-Specific Data: Task Data.}
The maximum number of training instances for all languages is again provided in Table~\ref{tb:st}. As before, for 4 languages we simulate low-resource settings by taking only a sample of the available task data: for \textsc{fget} we work with 200, 2K or 20K training instances which roughly correspond to training regimes of different data availability, while we select 300,\footnote{With a smaller number of instances (e.g., 100), \textsc{ner} and \textsc{mget} model training was unstable and resulted in near-zero performance across multiple runs.} 1K, and 10K sentences for \textsc{ner} and \textsc{mget}. Again, for the remaining 12 languages, we use all the available data to run the experiments. We adopt existing data splits into training, development, and test portions for \textsc{mtag} \cite{cotterell2017crosslingual}, and random splits for \textsc{fget} \cite{Heinzerling2018BPEmbTP,Zhu:2019arxiv} and \textsc{ner} \cite{pan2017cross}.


A large number of data points for scarcity simulations allow us to trace how performance on the three tasks varies in relation to the availability of WE data versus task data, and what data source is more important for the final performance.

\vspace{1.4mm}
\noindent \textbf{Embedding Training Setup.}
When training our subword-informed representations, we argue that keeping hyper-parameters fixed across different data points will possibly result in underfitting for larger data sizes or overfitting for smaller data sizes. Therefore, we split data points into three groups: $[10K, 50K]$ (G1), $(50K, 500K]$ (G2) and $(500K, 5M]$ (G3), and  use the same hyper-parameters for word embedding training within the same group. For G1, we train with batch size $32$ for $60$ epochs and set the minimum word frequency threshold to $2$. For G2 the values are: $128/30/3$, and $512/15/5$ for G3. This way, we ensure that 1) the difference of the absolute data sizes can be compared within the same data group, and 2) for the corresponding data points in different groups ($10K$, $100K$, $1M$) the sample efficiency can be compared, as the models trained on these data points undergo roughly the same number of updates.\footnote{We train \texttt{fastText} and skip-gram from \texttt{word2vec} with the same number of epochs that is used to train our subword-informed models on the corresponding data points.}

\vspace{1.4mm}
\noindent \textbf{Task-Specific Training Setup.}
For \textsc{fget}, we use the dataset of \citet{Heinzerling2018BPEmbTP} obtained by mapping entity mentions from Wikidata \citep{42240} to their associated FIGER-based most notable entity type \cite{Ling:2012:FER:2900728.2900742}. For each language, we randomly sample up to 100k pairs of entity mentions with corresponding entity type and create random 60/20/20 train/dev/test splits. Our \textsc{fget} model is designed after the hierarchical architecture by \citet{Zhu:2019arxiv}. For each entity token, we first use our subword-informed model to obtain word representations, and then feed the token embedding sequence into a bidirectional LSTM with 2 hidden layers of size 512, followed by a projection layer which predicts the entity type.
We initialize our \textsc{fget} model with the pretrained subword model, and fine-tune it during training. With BERT, we input the entire entity mention and then use the representation of the special \texttt{[CLS]} token for classification. We train with early stopping, using Adam \cite{DBLP:journals/corr/KingmaB14} with default parameters across all languages.
As suggested by \citet{Wu:2019beto}, BERT hyper-parameters are more sensitive to smaller data sizes, so we tune them on the smallest data point with $200$ training instances. We follow \citet{Wu:2019beto} to select hyper-parameter candidates, i.e., $2e{-5}/3e{-5}/5e{-5}$ for learning rate, $16/32$ for batch size and triangular learning rate scheduler with first $10\%$ of batches as warm-up. We do an exhaustive search on four high resource languages: \textsc{en}, \textsc{tr}, \textsc{he}, \textsc{zh} and select the hyper-parameter combination with the best average score on the development sets.

For \textsc{mtag}, we evaluate on the multilingual morphological annotations provided by the Universal Dependencies project \citep{nivre2016universal} and adopt the experimental protocol of \citet{cotterell2017crosslingual}.
Specifically, we cast \textsc{mtag} as a sequence labeling task by treating the concatenation of all key-value pairs for a given word as the word's label.
As sequence labeling model, we train a bidirectional LSTM \citep{Hochreiter:1997:LSM:1246443.1246450,plank2016multilingual}, with two layers of size 1024 
and dropout 0.4, using early stopping on the development set. For experiments involving multilingual BERT, we fine-tune all of BERT's layers and feed the final layer into an LSTM before classification. The evaluation metric is per-label accuracy, i.e.,\ a word's morphological tag is either predicted correctly or not, and there is no partial credit for the correct prediction of only a subset of features.

We evaluate \textsc{ner} performance on WikiAnn \citep{pan2017cross}, a multilingual dataset which provides three-class entity type annotations which were automatically extracted from Wikipedia.
We train sequence labeling models using exactly the same architectures and hyper-parameters as in \textsc{mtag}, and report F1 scores.
As WikiAnn does not come with predefined train/dev/test sets, we create random 60/20/20 splits.

\section{Results and Discussion}
\label{s:results}
\begin{figure*}[!t]
	\centering
    \includegraphics[width=.99\textwidth]{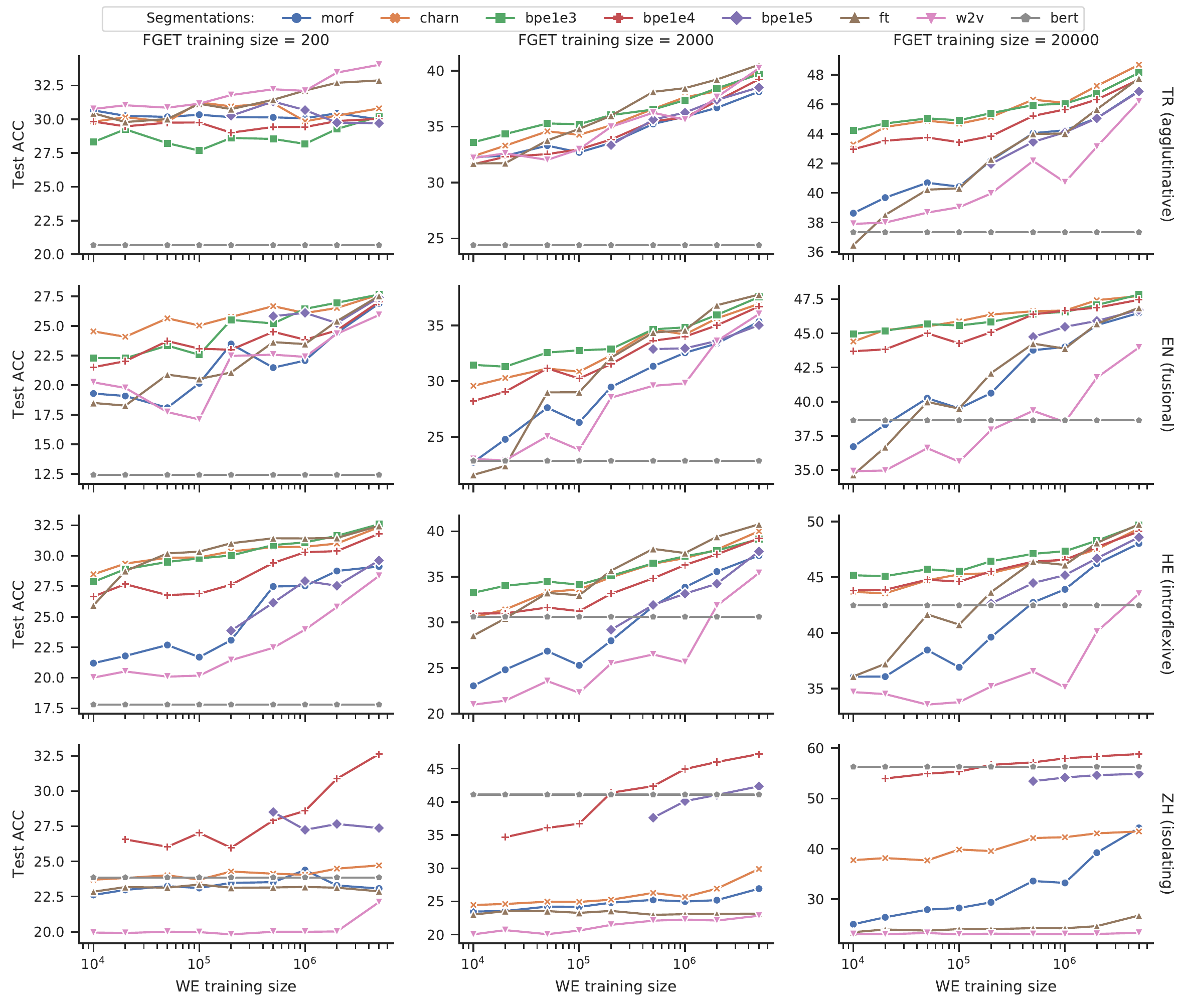}
    \vspace{-2mm}
    \caption{Test performance (accuracy) in the \textsc{fget} task for different segmentation methods in data scarcity simulation experiments across 4 languages representing 4 broad morphological types, averaged over 5 runs. Some data points with Chinese (\textsc{zh}) are not shown as in those cases the subword model is reduced to single characters only.} \label{fig:et_test}
    \vspace{-3mm}
\end{figure*}

\begin{figure*}[!t]
	\centering
    \includegraphics[width=0.99\textwidth]{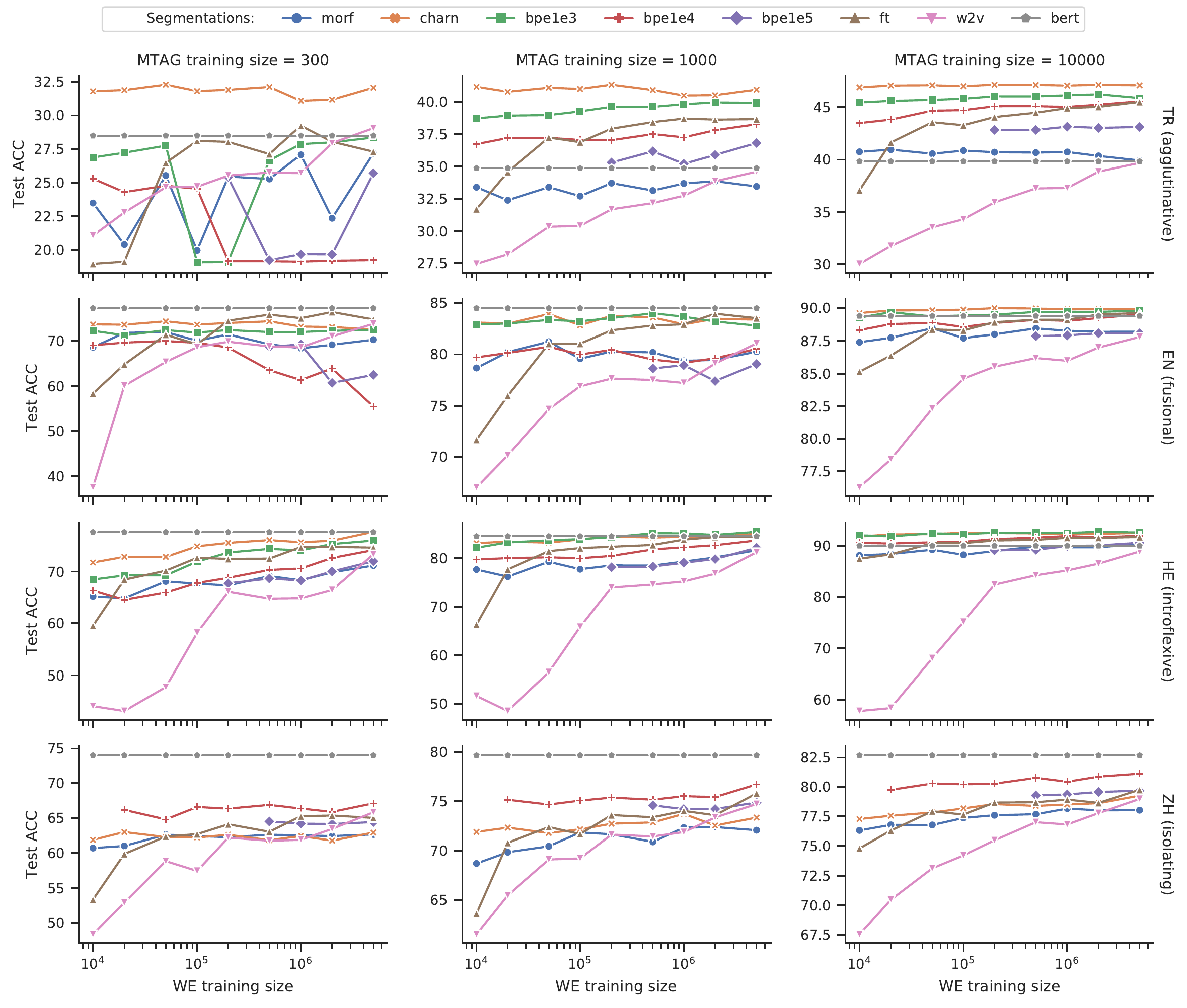}
    \vspace{-2mm}
    \caption{Test performance in the \textsc{mtag} task in data scarcity simulation experiments.} \label{fig:ner_test}
    \label{fig:mtag_test}
    \vspace{-3mm}
\end{figure*}

\begin{figure*}[t]
	\centering
    \includegraphics[width=0.99\textwidth]{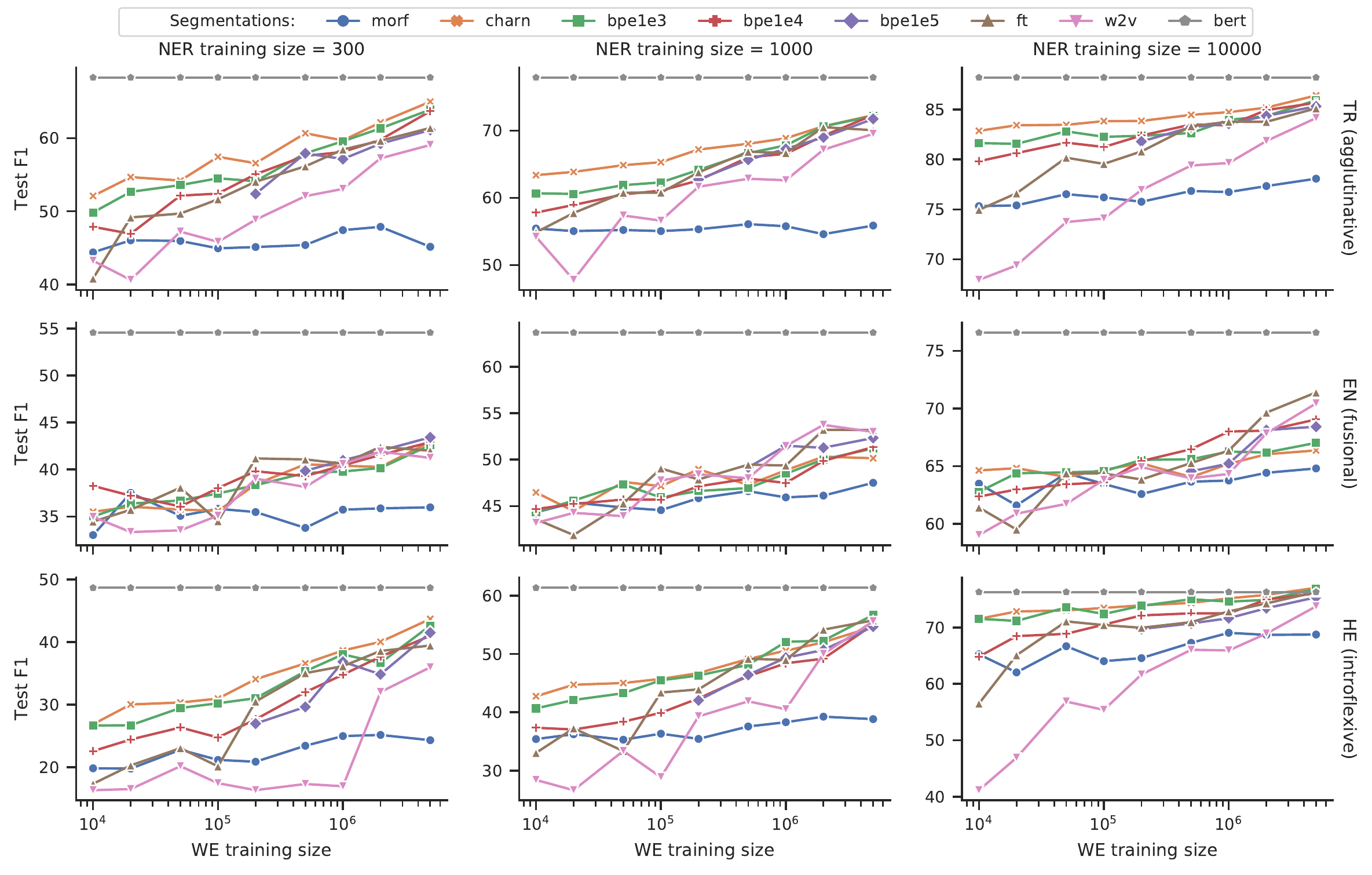}
    \vspace{-2mm}
    \caption{Test performance (F1 score) in the \textsc{ner} task in data scarcity simulation experiments. We do not show results for Chinese as the annotations of the Chinese \textsc{ner} are provided only on the character-level and thus impede experimentation with most of the subword-informed methods used in our evaluation.} \label{fig:ner_test}
    \vspace{-4mm}
\end{figure*}

\begin{table*}[t]
	\centering
    \def\arraystretch{0.95}
    {\footnotesize
	\begin{tabularx}{\textwidth}{lr XXXXX XXXXX X X}
		\multicolumn{2}{r}{}  & \multicolumn{5}{c}{Agglutinative} & \multicolumn{5}{c}{Fusional} & \multicolumn{1}{c}{Intro} & \multicolumn{1}{c}{Isolat} \\ \cmidrule(lr){3-7} \cmidrule(lr){8-12} \cmidrule(lr){13-13} \cmidrule(lr){14-14}
		
		\multicolumn{2}{r}{} & 
		\textsc{bm} & \textsc{bxr} & \textsc{myv} & \textsc{te} & \textsc{zu} &
		\textsc{fo} & \textsc{ga} & \textsc{got} & \textsc{mt} & \textsc{rue} &
        \textsc{am} &
		\textsc{yo} \\ \cmidrule(lr){3-14}
		
		\multirow{8}{*}{\textsc{\textsc{fget}}} 
& \texttt{morf}   & 52.43     & 52.47     & 79.11     & 57.79     & 53.00     & 54.43     & 50.77     & 29.90     & 49.48     & 50.38     & 41.82           & 83.43 \\
& \texttt{charn}  &{\bf 56.09}& 57.33     & 81.69     & 58.83     &{\bf 56.34}&{\bf 58.44}&{\bf 52.62}&{\bf 34.02}&{\bf 54.46}&{\bf 58.59}& 45.65           & 84.85 \\
& \texttt{bpe1e3} & 53.61     & 51.30     & 81.13     & 58.73     & 55.41     & 56.04     & 50.74     & 31.55     & 52.79     & 55.57     & {\bf 47.99}&{\bf 85.22} \\
& \texttt{bpe1e4} & 54.20     & 53.81     &{\bf 81.93}&{\bf 59.24}& 55.67     & 56.67     & 51.47     & 26.39     & 52.15     & 54.81     & 47.05           & 84.42 \\
& \texttt{bpe1e5} & -         & 53.80     & 80.00     & 58.13     & -         & 56.31     & 51.52     & -         & 51.52     & 52.52     & 44.74           & 83.39 \\
& \texttt{ft}     & 51.91     &{\bf 57.96}& 81.05     & 57.79     & 52.62     & 53.74     & 49.67     & 31.96     & 53.95     & 53.64     & 44.80           & 83.71 \\
& \texttt{w2v}    & 52.28     & 42.19     & 76.86     & 56.99     & 52.95     & 53.07     & 49.07     & 24.53     & 46.61     & 47.36     & 36.81           & 82.56 \\
& \texttt{bert}   & -         & -         & -         & 49.20     & -         & -         & 47.09     & -         & -         & -         & -               & 81.76 \\ \cmidrule(lr){3-14}

		\multirow{7}{*}{\textsc{\textsc{ner}}} 
& \texttt{morf}   & 73.29     & 76.58     & 83.40     & 77.01     & 65.22     & 84.29     & 86.94     & 59.49     & 74.37     & 81.87     & 66.67     & 90.01\\
& \texttt{charn}  &{\bf 83.02}&{\bf 81.59}&{\bf 93.22}&{\bf 88.23}&{\bf 74.47}&{\bf 91.08}& 88.95     &{\bf 84.99}&{\bf 83.56}    &{\bf 88.70}&{\bf 72.92}& 94.68\\
& \texttt{bpe1e3} & 77.22     & 79.33     & 89.00     & 85.82     & 71.91     & 89.73     & 89.18     & 81.03     & 81.63     & 85.30     & 70.84 & 92.35\\
& \texttt{bpe1e4} & 76.43     & 79.73     & 89.00     & 85.44     & 65.22     & 89.25     & 88.48     & 70.59     & 80.26     & 86.39     & 64.07     & 92.47\\
& \texttt{bpe1e5} & -         & 80.65     & 89.36     & 84.02     & -         & 88.66     &{\bf 89.48}& -         & 81.64     & 86.12     & 68.95     & 93.07 \\
& \texttt{ft}     & 73.29     & 79.81     & 88.57     & 86.88     & 58.16     & 89.48     & 89.18     & 58.16     & 81.64     & 83.54     & 68.29     & 92.58\\
& \texttt{w2v}    & 69.57     & 79.66     & 87.50     & 82.97     & 62.37     & 87.81     & 87.99     & 58.56     & 79.43     & 84.21     & 61.37     & 89.57\\
& \texttt{bert}   & -         & -         & -         & 82.31     & -         & -         & 88.45     & -         & -         & -         & -         &{\bf 95.53} \\
		\bottomrule       
	\end{tabularx}}%
    \vspace{-2mm}
     \caption{Test accuracy for \textsc{fget} and test F1 score \textsc{ner} for the $12$ low-resource test languages. The results are obtained by training on the full WE data (except for BERT) and the full task data of the corresponding languages.} \label{tb:results} 
     \vspace{-3mm}
\end{table*}

\begin{table}[t]
	\centering
    \def\arraystretch{0.95}
    {\footnotesize
	\begin{tabularx}{\columnwidth}{r XXX}
		\multicolumn{1}{r}{}  & Agg & Fus & Int  \\ \cmidrule(lr){2-2} \cmidrule(lr){3-3} \cmidrule(lr){4-4} 
		
		\multicolumn{1}{r}{} & \textsc{te} & \textsc{got} & \textsc{mt} \\ \cmidrule(lr){2-4}
		
		\texttt{morf}    & 87.79       & 76.94        & 92.80       \\
		\texttt{charn}   & 90.29       & {\bf 85.71}  & {\bf 94.39} \\
		\texttt{bpe1e3}  & 90.01       & 82.07        & 94.16       \\
		\texttt{bpe1e4}  & 87.79       & 83.28        & 92.82       \\
		\texttt{bpe1e5}  & 87.10       & -            & 92.51       \\
		\texttt{ft}      & {\bf 90.85} & 76.50        & 93.91       \\
		\texttt{w2v}     & 85.71       & 29.63        & 90.43       \\
		\texttt{bert}    & 87.45       & -            & -           \\ 		
		
		\bottomrule       
	\end{tabularx}}%
    \vspace{-2mm}
     \caption{Test accuracy for \textsc{mtag} for low-resource languages from UD where train/dev/test sets are available.} \label{tb:results_mtag}
     \vspace{-3mm}
\end{table}

\begin{figure*}[t]
	\centering
    \includegraphics[width=0.99\columnwidth, trim={2.5cm 1 2 0}, clip]{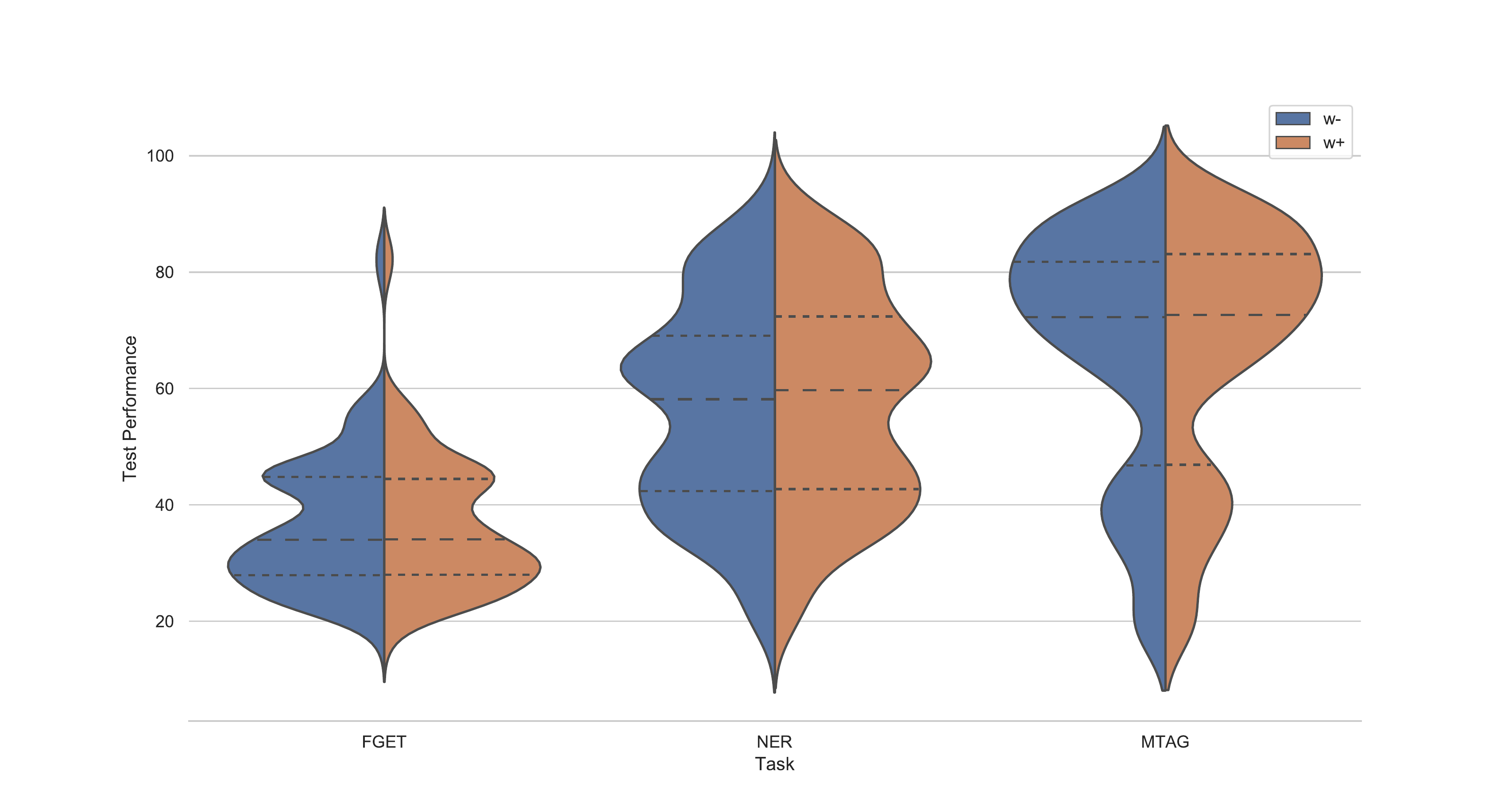}
    \includegraphics[width=0.99\columnwidth, trim={2.5cm 1 2 0}, clip]{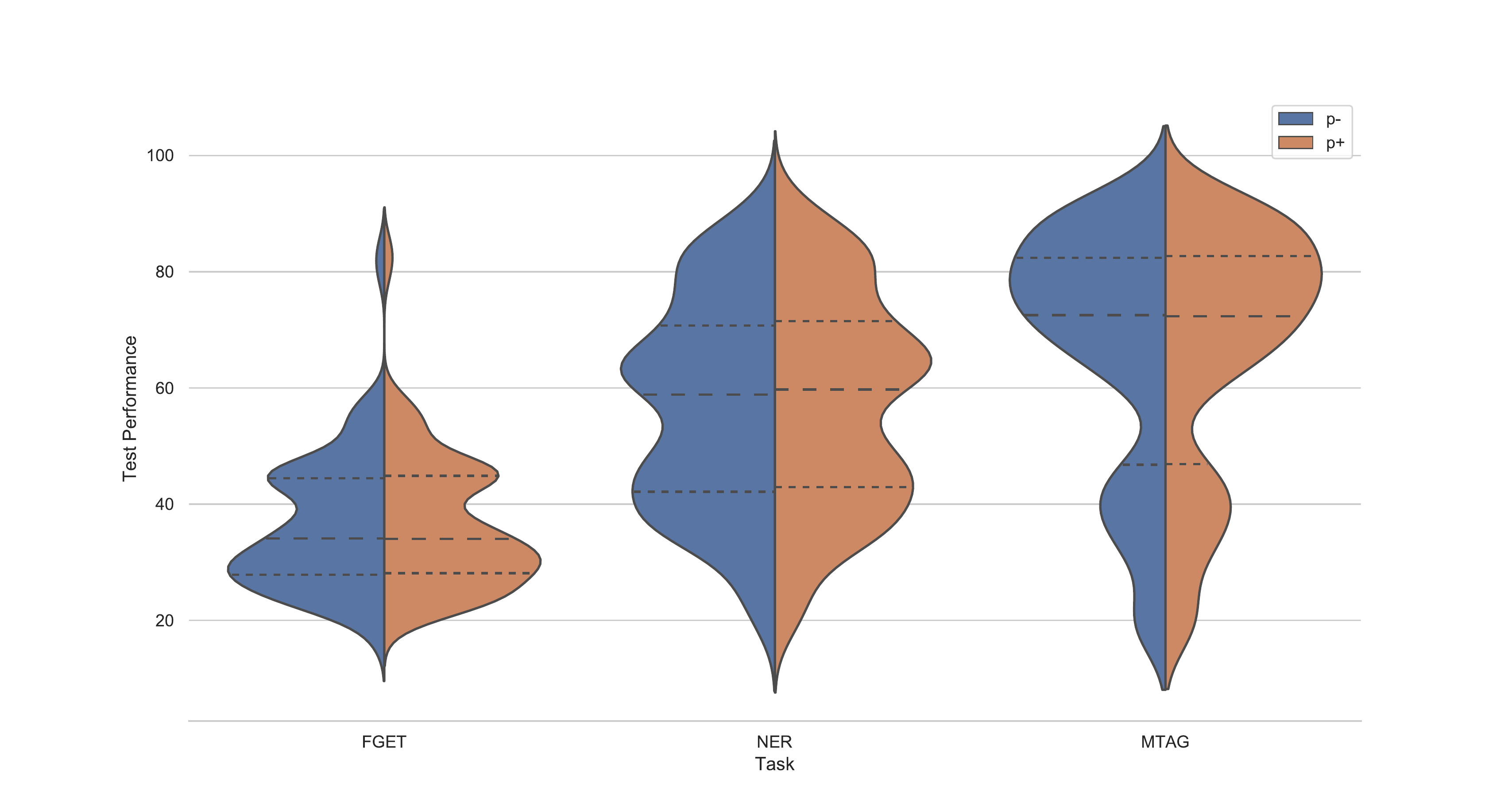}
    \vspace{-2mm}
    \caption{Comparisons of configurations with and without word token ({\it w-, w+}) and the position embedding ({\it p-, p+}). The results are obtained by collecting all data points in the data scarcity simulation for four high resource languages and the other 12 languages with both full WE data and task data.} \label{fig:wwpp}
    \vspace{-3mm}
\end{figure*}


Results for data scarcity simulation experiments are summarized in Figures~\ref{fig:et_test}-\ref{fig:ner_test}, while the results on the remaining 12 languages for all three tasks are provided in Tables~\ref{tb:results}-\ref{tb:results_mtag}, with the best results among different configurations of subword-informed methods reported. As the first main finding, the results show that subword-informed architectures substantially outperform the subword-agnostic skip-gram \textsc{w2v} baseline, and the gaps are in some cases very large: e.g., see the results in Figures~\ref{fig:et_test}-\ref{fig:ner_test} for the settings with extremely scarce WE data. These scores verify the importance of subword-level knowledge for low-resource setups.

\vspace{1.4mm}
\noindent \textbf{Simulating Data Scarcity.}
Another general finding concerns the importance of WE data versus task data. The simulation results in Figure \ref{fig:et_test}-\ref{fig:ner_test} suggest that both types of data are instrumental to improved task performance: This finding is universal as we observe similar behaviors across tasks and across different languages. While WE data is important, considerably larger gains are achieved by collecting more task data: e.g.,\ see the large gains in \textsc{fget} when training on 200 versus 2K entity mentions. In summary, both types of data scarcity decrease performance, but the impact of scarce task data seems more pronounced. Collecting more WE data when dealing with scarce task data leads to larger gains in the \textsc{fget} task compared to \textsc{mtag} or \textsc{ner}.

While subword models are generally better than the baselines across different data points, less aggressive segmentation models and token-based models close the gap very quickly when increasing WE data, which is in line with the findings of \citet{Zhu:2019arxiv}, where \texttt{morf} eventually prevails in this task with abundant WE data. This again verifies the usefulness of subword-level knowledge for low-(WE) data regimes. Similar trends emerge in terms of task data, but the advantage of subword models seems more salient with more task data. The underlying task architectures start making use of subword features more effectively: this shows that subword-level knowledge is particularly useful for the three chosen morphological tasks.



\vspace{1.4mm}
\noindent \textbf{Performance of BERT.}
An interesting analysis regarding the (relative) performance of pretrained multilingual BERT model emerges from Figures~\ref{fig:et_test}-\ref{fig:ner_test}. Fine-tuned BERT displays much stronger performance in low-resources settings for the \textsc{mtag} and \textsc{ner} tasks than for the \textsc{fget} task (e.g., compare the sub-figures in the first columns of the corresponding figures). The explanation of \textsc{mtag} and \textsc{ner} performance is intuitive. A pretrained BERT model encodes a massive amount of background knowledge available during its (multilingual) pretraining. However, as we supplement other subword-informed representation learning methods with more data for training the embeddings, the gap gets smaller until it almost completely vanishes: other methods now also get to see some of the distributional information which BERT already consumed in its pretraining.

BERT performance on \textsc{fget} versus \textsc{mtag} and \textsc{ner} concerns the very nature of the tasks at hand. The input data for \textsc{fget} consist mostly of 2-4 word tokens (i.e., entity mentions), while \textsc{mtag} and \textsc{ner} operate on full sentences as input. Since BERT has been pretrained on sentences, this setting is a natural fit and makes fine-tuning to these tasks easier: BERT already provides a sort of ``contextual subword composition function''. This stands in contrast with the other subword-informed approaches. There, we might have good non-contextual subword embeddings and a pretrained ``non-contextual'' composition function, but we have to learn how to effectively leverage the context for the task at hand (i.e., by running an LSTM over the subword-informed token representations) from scratch.\footnote{Another factor at play is multilingual BERT's limited vocabulary size (100K WordPiece symbols), leaving on average a bit under 1K symbols per language. Due to the different sizes of Wikipedias used for pretraining BERT, some languages might even be represented with far fewer than 1K vocabulary entries, thereby limiting the effective language-specific model capacity. Therefore, it is not that surprising that monolingual subword-informed representations gradually surpass BERT as more language-specific WE data becomes available. This finding is also supported by the results reported in Table~\ref{tb:results}.}


\vspace{1.4mm}
\noindent \textbf{Truly Low-Resource Languages.}
The results on the 12 test languages in Table \ref{tb:results}-\ref{tb:results_mtag} suggest that subword-informed models are better than the baselines in most cases: this validates the initial findings from the simulation experiments. That is, leveraging subword information is important for WE induction as well as for task-specific training. The gains 
with subword methods become larger for languages with fewer WE data (e.g., \textsc{zu}, \textsc{bm}, \textsc{got}); this is again consistent with the previously reported simulation experiments. 


\vspace{1.4mm}
\noindent \paragraph{Tasks, Language Types, Subword Configurations.}
The results further suggest that the optimal configuration indeed varies across different tasks and language types, and therefore it is required to carefully tune the configuration to reach improved performance. For instance, as agglutinative languages have different granularities of morphological complexity, it is not even possible to isolate a single optimal segmentation method within this language type. Overall, the segmentation based \texttt{charn} followed by BPE emerge as most robust choices across all languages and tasks. However, \texttt{charn} has the largest number of parameters and is slower to train compared to other segmentations, and in case of BPE its number of merge operations must be tuned to yield competitive scores.

While we do not see extremely clear patterns from the results in relation to particular language types, the scores suggest that for agglutinative and fusional languages a hybrid segmentation such as \texttt{charn} or a moderate one (\texttt{bpe1e4}, \texttt{bpe1e5}) is a good choice. For introflexive and isolating languages, more aggressive segmentations seem to be also competitive in \textsc{fget} and \textsc{mtag}, while \texttt{bpe1e4} being very effective for \textsc{zh}, and \texttt{charn} (again) and \texttt{bpe1e5} seems to be preferred in \textsc{ner}.

Apart from segmentation methods, we also analyzed the effect of word token embeddings ({\it w+}) and position embeddings ({\it p+}) in the subword-informed learning framework \cite{Zhu:2019arxiv} (see before Table~\ref{tb:sub_model} in \S\ref{s:methodology}), shown in Figure \ref{fig:wwpp}.
\textsc{ner} can clearly benefit from both {\it w+} and {\it p+} and {\it w+} is also useful for \textsc{mtag}.
However, for other tasks, the fluctuations between configurations are minimal once the segmentation has been fixed, which suggests that the most critical component is indeed the chosen segmentation method: this is why we have mostly focused on the analyses of the segmentation method and its impact on task performance in this work. Regarding the composition functions, as demonstrated in \citet{Zhu:2019arxiv}, more complex composition functions do not necessarily yield superior results in a range of downstream tasks. We therefore leave the exploration of more sophisticated composition functions for future work.

\section{Conclusions and Future Work}
\label{s:conclusion}
We have presented an empirical study focused on the importance of subword-informed word representation architectures for truly low-resource languages. Our experiments on three diverse morphological tasks with 16 typologically diverse languages of varying degrees of data scarcity have validated that subword-level knowledge is indeed crucial for improved task performance in such low-data setups. The large amount of results reported in this work has enabled comparisons of different subword-informed methods in relation to multiple aspects such as the degree of data scarcity (both in terms of embedding training data and task-specific annotated data), the task at hand, the actual language, as well as the methods' internal design (e.g.\ the choice of the segmentation method). Our results have demonstrated that all these aspects must be considered in order to identify an optimal subword-informed representation architecture for a particular use case, that is, for a particular language (type), task, and data availability. However, similar paterns emerge: e.g., resorting to a segmentation method based on character n-grams seems most robust across the three tasks and across languages, although there are clear outliers. In future work, we will extend our focus to other target languages, including the ones with very limited \cite{adams2017cross} or non-existent digital footprint.

\section*{Acknowledgments}
This work is supported by the ERC Consolidator Grant LEXICAL: Lexical Acquisition Across Languages (no 648909) and the Klaus Tschira Foundation, Heidelberg, Germany. We thank the three anonymous reviewers for their helpful suggestions.

\bibliography{references}
\bibliographystyle{acl_natbib}

\end{document}